\documentclass{article}

\usepackage{microtype}
\usepackage{graphicx}
\usepackage{booktabs} 
\usepackage{caption}

\usepackage{hyperref}

\usepackage[accepted]{tom2023}

\usepackage{amsmath}
\usepackage{amssymb}
\usepackage{mathtools}
\usepackage{amsthm}

\usepackage[capitalize,noabbrev]{cleveref}

\theoremstyle{plain}

\theoremstyle{definition}

\theoremstyle{remark}

\usepackage[textsize=tiny]{todonotes}

\icmltitlerunning{Attention Schema in Neural Agents}

\begin{document}

\twocolumn[
\icmltitle{Attention Schema in Neural Agents}




\begin{icmlauthorlist}
\icmlauthor{Dianbo Liu}{ml}
\icmlauthor{Samuele Bolotta}{ml,udm}
\icmlauthor{He Zhu}{ml,mc}
\icmlauthor{Yoshua Bengio}{ml,udm,cf}
\icmlauthor{Guillaume Dumas}{ml,udm,cf}

\end{icmlauthorlist}

\icmlaffiliation{ml}{Mila-Quebec AI institute, Montreal, Canada}
\icmlaffiliation{udm}{University of Montreal, Montreal, Canada}
\icmlaffiliation{mc}{McGill University, Montreal, Canada}
\icmlaffiliation{cf}{CIFAR, Canada}

\icmlcorrespondingauthor{Guillaume Dumas}{guillaume.dumas@umontreal.ca}
\icmlcorrespondingauthor{Dianbo Liu}{dianbo.liu@mila.quebec}
\icmlkeywords{Theory of Mind, ToM}

\vskip 0.3in
]



\printAffiliationsAndNotice{}  

\begin{abstract}
Attention has become a common ingredient in deep learning architectures. It adds a dynamical selection of information on top of the static selection of information supported by weights. In the same way, we can imagine a higher-order informational filter built on top of attention: an \emph{Attention Schema (AS)}, namely, a descriptive and predictive model of attention. In cognitive neuroscience, Attention Schema Theory (AST) supports this idea of distinguishing attention from AS. A strong prediction of this theory is that an agent can use its own AS to also infer the states of other agents' attention and consequently enhance coordination with other agents. As such, multi-agent reinforcement learning would be an ideal setting to experimentally test the validity of AST. We explore different ways in which attention and AS interact with each other. Our preliminary results indicate that agents that implement the AS as a \textit{recurrent internal control} achieve the best performance. In general, these exploratory experiments suggest that equipping artificial agents with a model of attention can enhance their social intelligence.
\end{abstract}

\section{Introduction}

In deep learning, attention can be understood as a dynamical control of information flow \cite{mittal2020learning}. In the last decade, the scope of attention mechanisms has grown from the first implementation in RNNSearch \cite{bahdanau2014neural} to the recent large-scale models currently dominating natural language processing and text-to-image generation \cite{santana2021neural}. Transformers have particularly demonstrated how \emph{attention may be all we need}, from sequence learning \cite{vaswani2017attention} to visual processing \cite{dosovitskiy2020image} and time series forecasting \cite{lim2021temporal}.

\begin{figure*}[t]
\vspace{0.5cm}
    \centering
    \includegraphics[width=1.0\textwidth]{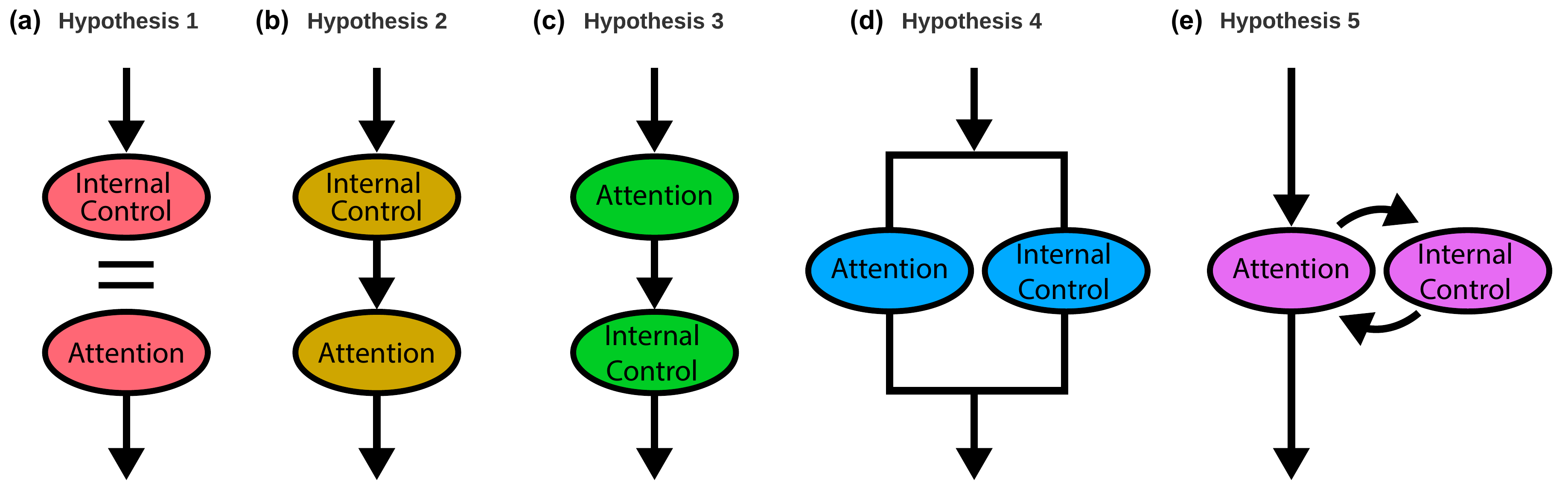}
    \caption{Hypotheses about the relationship between attention and internal control of attention. Adapted from \cite{webb2015attention}.}
    \label{fig:hypothesis}
\end{figure*}

Although transformers proposed a general-purpose architecture in which inductive biases shaping the flow of information are learned from the data itself, there is still the possibility to add a third-order control of the information flow. Indeed, the weights between neurons offer a first-order static control of the flow of information, whereas the attention then introduces a second-order dynamical control of the flow of information. The key reason why a higher-order filter on top of attention seems a promising idea for deep learning comes from control engineering: a good controller contains a model of the item being controlled \cite{conant1970}. More specifically, a descriptive and predictive model of attention (referred to as "Internal Control" in Figure \ref{fig:hypothesis}) could help the dynamical control of attention and therefore maximize the efficiency with which resources are strategically devoted to different elements of an ever-changing environment \cite{graziano2017}. In fact, the performance of an artificial agent in solving a simple sensorimotor task has been shown to be greatly enhanced by a simple model of attention, but greatly reduced when such a model was not available \cite{wilterson2021}. 

In this regard, attention schema theory (AST) is a neuroscientific theory that postulates that the human brain, and possibly the brain of other animals, constructs a model of attention: an attention schema \cite{graziano2015}. Such an attention schema is a coherent set of information that represents the basic properties of attention and its state dynamics (i.e., how it evolves in time). This internal model of attention allows to predict where attention can be directed so that it stays within the desired range of each object. Specifically, there are two predictions of this theory that are investigated in this paper. The first prediction is that, without an attention schema, attention is still possible, but it suffers deficits in control and thus leads to worse performance. The second prediction is that an attention schema is also useful to model the attention of other agents \textemdash since the machinery that computes information about the attention of other people is the same machinery that computes information about our own attention \cite{graziano2011}.


Building a rich, flexible, and coherent model of attention is a complex task for AI, especially because to express all of its potential, language and higher-order cognition must have access to it \cite{wilterson2021}. In this paper, we are not trying to solve the entire problem but rather the foundation for future research on this topic. Specifically, we want to start tackling the following question: Can equipping agents with self-monitoring capabilities boost their ability to intelligently control and deploy their limited processing resources? To investigate this, we implement internal control as a recurrent network and attention as a key-value attention mechanism; we then compare five different possible relationships between attention and its internal control, as suggested in the cognitive science literature (Figure \ref{fig:hypothesis}) \cite{graziano2015attention}. A strong prediction of AST is that an agent can use attention and its own internal control to also infer the states of the attention of other agents. As a consequence, this is expected to enhance coordination with other agents. Therefore, we test the five different hypotheses in multi-agent reinforcement learning environments in which cooperation among agents is important. Our results suggest that higher performance is achieved by agents modeled according to the fifth hypothesis, which corresponds to AST and postulates the existence of an internal control serving as an inner regulation of the attention mechanism.

\section{Internal control and attention}


Potentially, model-based reinforcement learning offers numerous benefits over model-free reinforcement learning. Real-world samples are expensive, but learning a model of the environment can improve data efficiency, prompt targeted exploration, and promote better asymptotic convergence, which is crucial for scaling reinforcement learning (RL) to real-world applications \cite{moerland2021}. This paper is well aligned with this intuition: instead of simply focusing attention on various elements of the environment and then reinforcing those actions in which a higher reward was obtained, a simple model of attention can be used to exert greater control on behavior. In other words, our goal is to leverage the relationship between external stimuli, their representation in the brain of the agent (coordinated by attention), and the successful control of behavior (coordinated by internal control). To unravel the details of such a relationship, we set out five hypotheses about the relationship between attention and its internal control, as illustrated in Figure \ref{fig:hypothesis}.


\subsection{Internal control and attention are the same}

Attention and its internal control are the same. The purpose of this setup is to act as a baseline. In humans, there is evidence of cases where attention unfolds without its internal control being active, but this seems to correlate with poorer performance \cite{graziano2015}.

\subsection{Internal control precedes attention}

The system can only focus its attention on what has been modeled. In other words, the external stimulus is not attended immediately. There are two problems with this scenario: a) attention seems to be necessary to bind together the different components of a representation, which makes it impossible for the representation of a stimulus to be bound to the information contained in its internal control \cite{treisman1980} b) without attention, the representation of a stimulus has lower signal strength and is less likely to have an influence on the policy \cite{graziano2015}.

\subsection{Attention precedes internal control}

The system can only model what is under the focus of attention. This hypothesis is close to what AST proposes, except for one aspect: internal control is not modeling attention in a recurrent way. This makes the system less flexible, and therefore less efficient in keeping up with ever-changing environments.

\subsection{Internal control and attention are independent processes}

The system simultaneously models the input and focuses attention on some sub-parts of the input. Both processes occur in a parallel fashion. Considering how closely related these two components are in humans, it is unlikely that implementing them as independent processes can lead to good performance.

\subsection{Internal control of attention: the attention schema}

This hypothesis is the most accurate representation of AST. The system models attention and can control it. The internal control corresponds to the attention schema: a coherent set of information that represents the basic properties of and changes in the state of attention in a dynamical fashion. This setup should offer the maximal potential for coordination between agents.

\section{Background and notation}

The whole Multi-Agent Reinforcement Learning (MARL) game is defined as $(N,S,A,P,R,O,\gamma)$. $N = \{1,...,N\}$ denotes the set of N $>$ 1 agents. $S_t=\{s_{1,t},s_{2,t},...,s_{N,t}\}$ denotes the state of each agent at time step t. $O_t=\{o_{1,t},o_{2,t},...,o_{N,t}\}$ denotes the partial observation of the environment that each agent receives as input at time step t. $O_t$ could be different from $S_t$. $A_{j_t}$ denotes the action of agent j at time step t . $P : S \times A \rightarrow \Delta(S)$ denotes the probability of state transition. $R^j :S\times A \times S \rightarrow R$ is the reward function that determines the immediate reward received by agent j for a transition from $(s;a)$ to $s'$ and $\gamma$ is the discount factor. 
\section{Methods}

The five hypotheses were implemented using PyTorch \cite{PyTorch_2019}, and tested on two tasks from two different MARL benchmarks: the GhostRun environment \cite{shuo2019maenvs} and the Multi-Agent Particle environment \cite{mordatch2017emergence}.


\subsection{Architectures}

The attention module, the internal control module, and their interactions are implemented according to hypotheses 1-5 as described in Section 2.
Attention modules are implemented as multi-head attention layers similar to the kind implemented by Transformers \cite{vaswani2017attention}; they dynamically select information from the observation space or, in the case of hypotheses 2 and 4, from the output of the internal control module. The internal control modules are implemented as recurrent neural networks (RNNs) with gated recurrent units (GRU), which take as input the time series signal or, in the case of hypotheses 2 and 4, the observation space; to make a fair comparison, the implementation of different hypotheses uses the same architectures of internal control of attention and other components that are shared by different hypotheses but without sharing parameters. All architectures are trained with proximal policy optimization (PPO). The formal definition of the architecture used to test each hypothesis follows. 
%

\textbf{Internal control module.} The internal control module is implemented as an RNN. The input varies between hypotheses; for each element of the input sequence $x_{j,t}$, each layer computes the following function:

            
            
            

       $\begin{array}{ll}
            r_t = \sigma(\bold{W} x_{j,t} + b+ \bold{W'} h_{(t-1)} + b') \\
            
            z_t = \sigma(\bold{\hat{W}} x_{j,t} + \hat{b} + \bold{\hat{W}'} h_{(t-1)} + \hat{b}') \\
            
            n_t = \tanh(\bold{\Bar{W}} x_{j,t} + \Bar{b} + r_t * (\bold{\Bar{W}'}h_{(t-1)}+ \Bar{b}')) \\
            
           h_t = (1 - z_t) * n_t + z_t * h_{(t-1)}
        \end{array}$
        
Where $h_t$ is the hidden state at time t, $x_{j,t}$ is the input at time t for agent j, $h_{(t-1)}$ is the hidden state of the layer at time t-1 and $r_t$, $z_t$ and $n_t$ are the reset, update, and new gates, and $\bold{W}$, $\bold{\hat{W}}$, $\bold{\Bar{W}}$ are the corresponding weights, respectively. $\sigma$ is the sigmoid function, and $*$ is the Hadamard product.

\textbf{Attention module}

The keys ($K$), values ($V$), and queries($Q$) vary across the hypotheses. Attention is computed as:

$Attention(Q, K, V) = softmax(\frac{QK^T}{\sqrt{d_k}})V$

where $d_k$ is the dimension of the vectors
\subsection{Hypothesis 1} 
Hypothesis 1 (Figure \ref{fig:hypothesis}a) postulates that attention and internal control are the same. Therefore, no internal control module is implemented. The partial visual observation of an agent $j$ at time step $t$ is rearranged into different patches as matrix $o^{pa}_{j,t}$, which has shape (number of patches $\times$ number of pixels per patch). The attention key is $K_j=\bold{W^k_j} o^{pa}_j$; the attention value is $V_j=\bold{W^v_j} o^{pa}_j$; the attention query $q_j$ is the vector $o^{pa}_j$ after taking the mean across the first dimension (number of patches). This design is motivated by the idea that an agent will decide where to focus based on the whole view it observes. Attention mechanisms in all other hypotheses follow the same strategy. The output vector of the attention module is $h1_{j,t}$ , which is a weigthed sum of values by attention scores, and is directly fed into the policy network in Hypothesis 1.

\begin{figure*}[]
    \centering
    \includegraphics[width=0.9\textwidth]{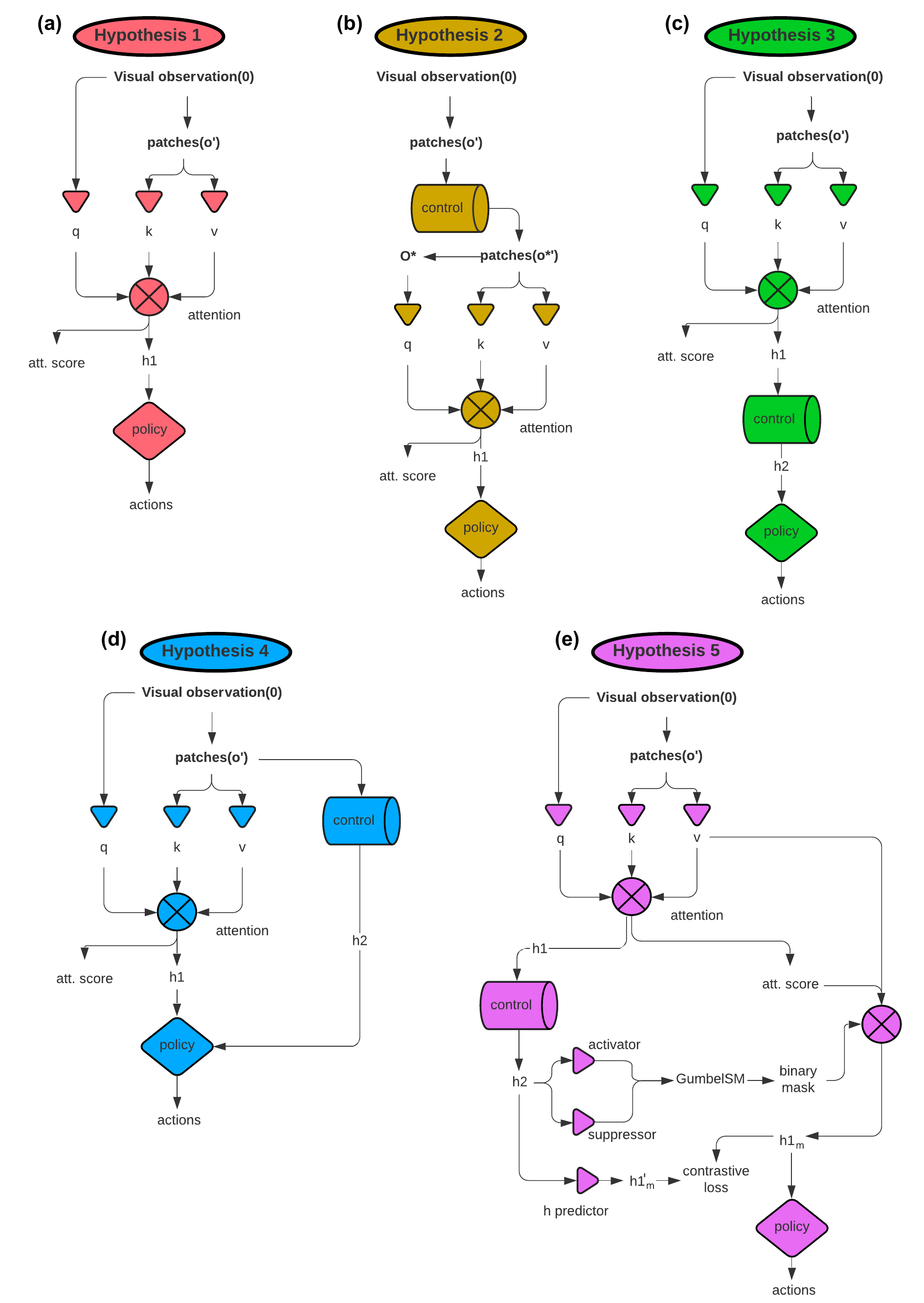}
    \caption{Architecture of the five hypotheses. "q" denotes the attention query; "k" denotes the attention key; "v" denotes the attention value; "h1" denotes the output of attention; "h2" denotes the output of internal control.}
\end{figure*}

\subsection{Hypothesis 2} 
 Hypothesis 2 (Figure \ref{fig:hypothesis}b) postulates that internal control precedes attention. The output vector $h2^{}_{j,t}$ of the internal control module is used as an input for attention, which outputs $h1^{}_{j,t}$ and feeds it directly into the policy network.

\begin{figure*}[H]
    \centering
    \includegraphics[width=0.9\textwidth]{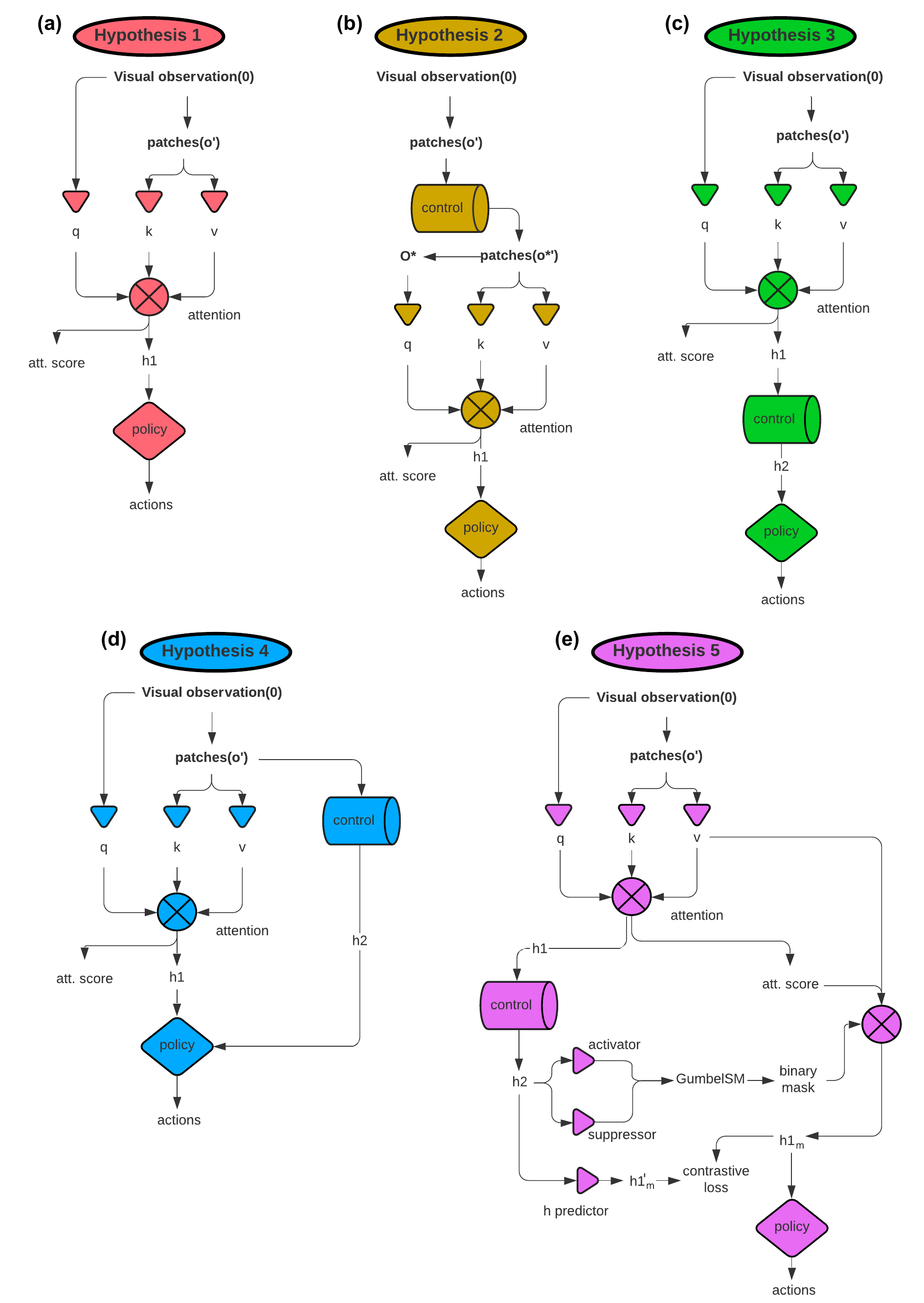}
    \caption{Architecture of the five hypotheses. "q" denotes the attention query; "k" denotes the attention key; "v" denotes the attention value; "h1" denotes the output of attention; "h2" denotes the output of internal control.}
\end{figure*}

\subsection{Hypothesis 3} 
 Hypothesis 3 (Figure \ref{fig:hypothesis}c) postulates that attention precedes internal control. The output vector $h1^{}_{j,t}$ of the attention module is used as input for the internal control module, which outputs $h2^{}_{j,t}$ and feeds it directly into the policy network.

\subsection{Hypothesis 4} 
 Hypothesis 4 (Figure \ref{fig:hypothesis}d) postulates that internal control and attention are independent processes. Attention is calculated in the observation space and outputs the vector $h1^{}_{j,t}$, which is directly fed into the policy network. In parallel, the internal module is applied to each element in the input sequence $o^{pa}_{j,t}$ and outputs the vector $h2^{}_{j,t}$, which is directly fed into the policy network.

\subsection{Hypothesis 5} 
 Hypothesis 5 (Figure \ref{fig:hypothesis}e) postulates that the internal control corresponds to the attention schema. The internal control module is implemented as a recurrent neural network (RNN). However, the output is not used as an input to the policy. Instead, the internal control module learns to predict the results of attention via contrastive loss. In the diagram, $h1^{}_{j,t}$ is the attention output vector and $h2^{}_{j,t}$ is the RNN output of the internal control. In this hypothesis $h2^{}_{j,t}$ is used to predict the final attention output $h1_m$ which is used as input into the policy network via a constrastive loss.

\[ContrastiveLoss=(h1'_{m}-h1_{m})^2\]

where $h1'_{m}$ is the vector predicted from $h2^{}_{j,t}$. In addition, $h2$ is used to generate a binary mask that is applied to the attention scores in the attention mechanism, resulting in: $AttScore'=AttScore \circ M_{att}$ where $M_{att} \in[0,1]^d_{att}$ is the binary mask and $\circ$ refers to element-wise multiplication. The mask is generated from a pair of MLPs inspired by a neural activator and suppressor and followed by a Gumbel softmax to binarize the value pairs \cite{jang2017}, where the activator and suppressor work together to generate a binary mask on attention scores, in which a "1" indicates that the corresponding attention score is allowed to be active and a "0" means that the attention is suppressed. 



\section{Experimental design and preliminary results}
The experiments were designed to explore which combination of attention and its internal control achieves the best performance in a multi-agent cooperative context. Therefore, we tested the five hypotheses in Figure \ref{fig:hypothesis} in multi-agent reinforcement learning (MARL) tasks where coordination is required to achieve a cooperative goal. Our preliminary results suggest that Hypothesis 5 shows marginal advantages over other hypotheses.

Next, we ablated and rearranged different components in the architecture of hypothesis five. While hypothesis 5.4 has already been described previously, these are the other four versions of hypothesis five:

- Hypothesis 5.1: There is no binary mask. In other words, there is no control.

- Hypothesis 5.2: The binary mask is applied to the action distribution predicted by the policy network. In other words, the control is on action and not on attention.

- Hypothesis 5.3: A binary mask is applied to the output of attention $h1_{m}$ rather than directly interfering with the attention mechanism. In other words, the control is on the output of attention and not on the attention itself.

- Hypothesis 5.5: It uses the prediction $h1'_{m}$ made by the internal control module as input into the policy network.

Architecture 5.4 is the one that best reflects what is proposed by attention schema theory because it is the only one in which control is exerted on attention. Our exploratory experiments suggest that hypotheses 5.4 and 5.5 show the best results.

\subsection{Tasks}

The experiments aim at investigating which combination of attention and its internal control achieves the best performance in a multi-agent cooperative context. Therefore, we tested the five hypotheses in Figure \ref{fig:hypothesis} in multi-agent reinforcement learning (MARL) tasks where coordination is required to achieve a cooperative goal.

We used two different environments where agents have a visual observation of their surroundings (RGB matrix of observed pixel values) and a discrete action space of movement (up/down/left/right).

\begin{figure}[h]
    \centering
    \includegraphics[width=0.5\textwidth]{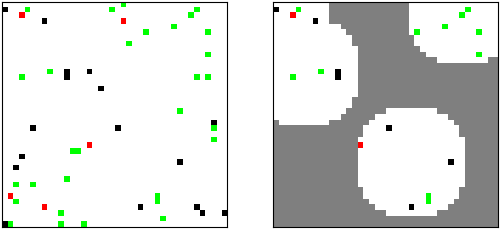}
    \caption{GhostRun environment: The agents need to keep away from ghosts (red pixels) as much as possible. The environment also contains trees (green pixels) and walls (black pixels). The right panel shows only the parts of the environment visible to the agents.}
    \label{fig:ghostrun}
\end{figure}

\paragraph{GhostRun Environment.} The GhostRun environment consists of multiple agents, each with a partial view of the environment. The space consists of ghosts, represented by red dots, trees, represented by green dots, and obstacles, represented by black dots. The ghosts move around randomly, whereas the trees and obstacles are stationary. The task for the team of cooperative agents is to escape from ghosts, or more quantitatively, to minimize the number of ghosts in each agent's partial observation of the environment. The reward received by each agent at each time step is the negative of the total number of ghosts in the view of all agents and a step cost of -1 for each step taken (Figure \ref{fig:ghostrun}). 

\begin{figure}[h]
    \centering
    \includegraphics[width=0.5\textwidth]{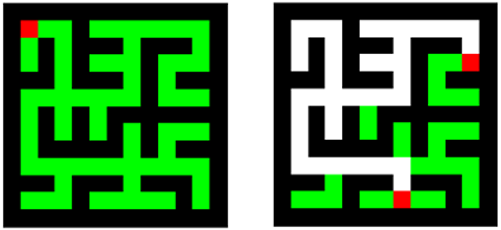}
    \caption{MazeCleaners environment: The agents (red pixels) need to clean the green parts of the maze as fast as possible while navigating between the walls (black pixels). The right side shows the maze on the left side after 30 epochs.}
    \label{fig:mazecleaners}
\end{figure}

\paragraph{MazeCleaners Environment.} In MazeCleaners, agents are moving in a maze and need to clean it cooperatively as quickly as possible. Agents receive a collective reward for each part of the maze cleaned up (Figure \ref{fig:mazecleaners}). Agents cannot walk through walls. The world size is 13 by 13, and the current scenario focuses on 2 agents. For OOD performance, the spawning location of the agents is random.

\begin{figure}
    \centering

    \includegraphics[width=0.24\linewidth]{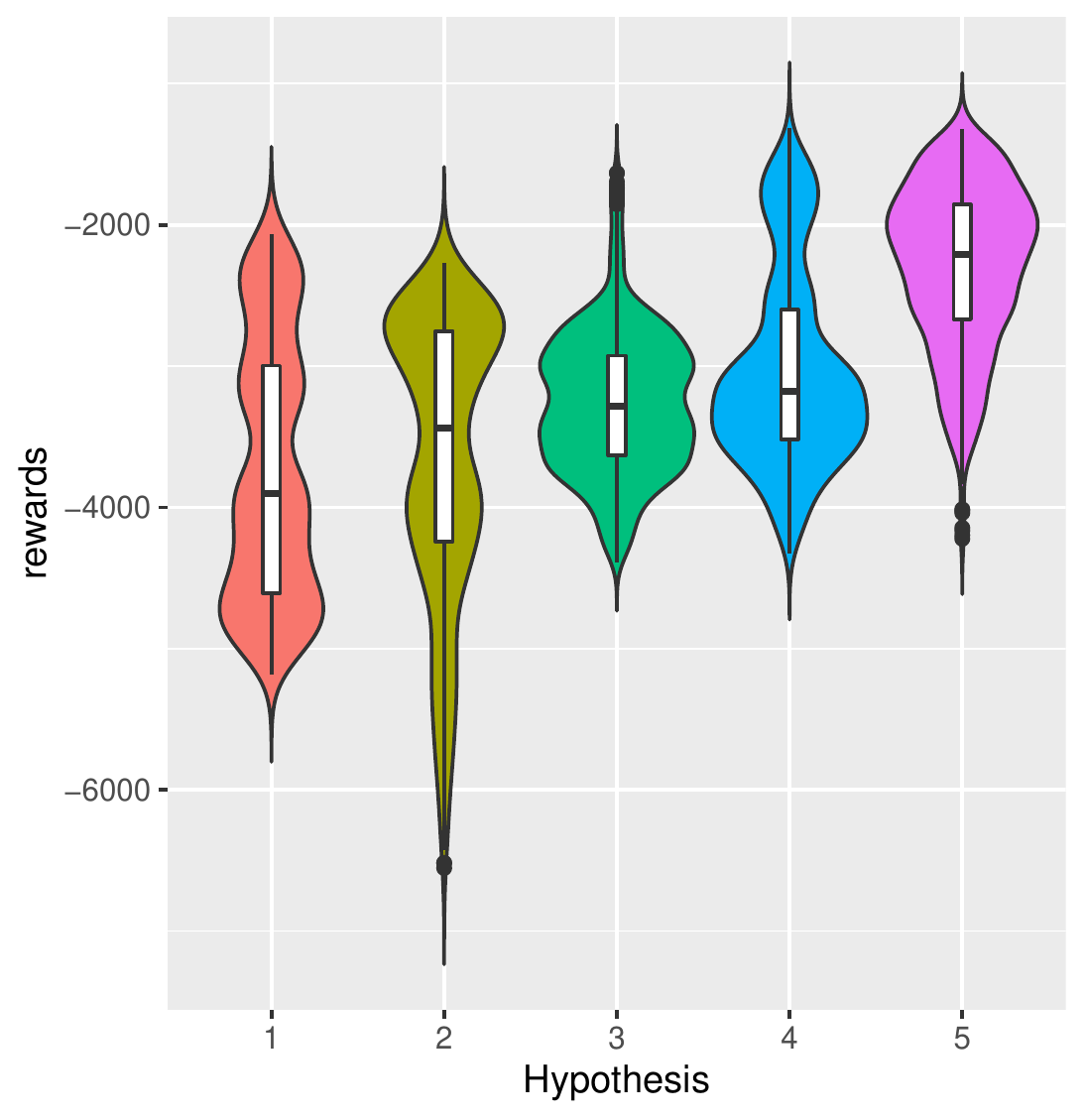}
    \includegraphics[width=0.24\linewidth]{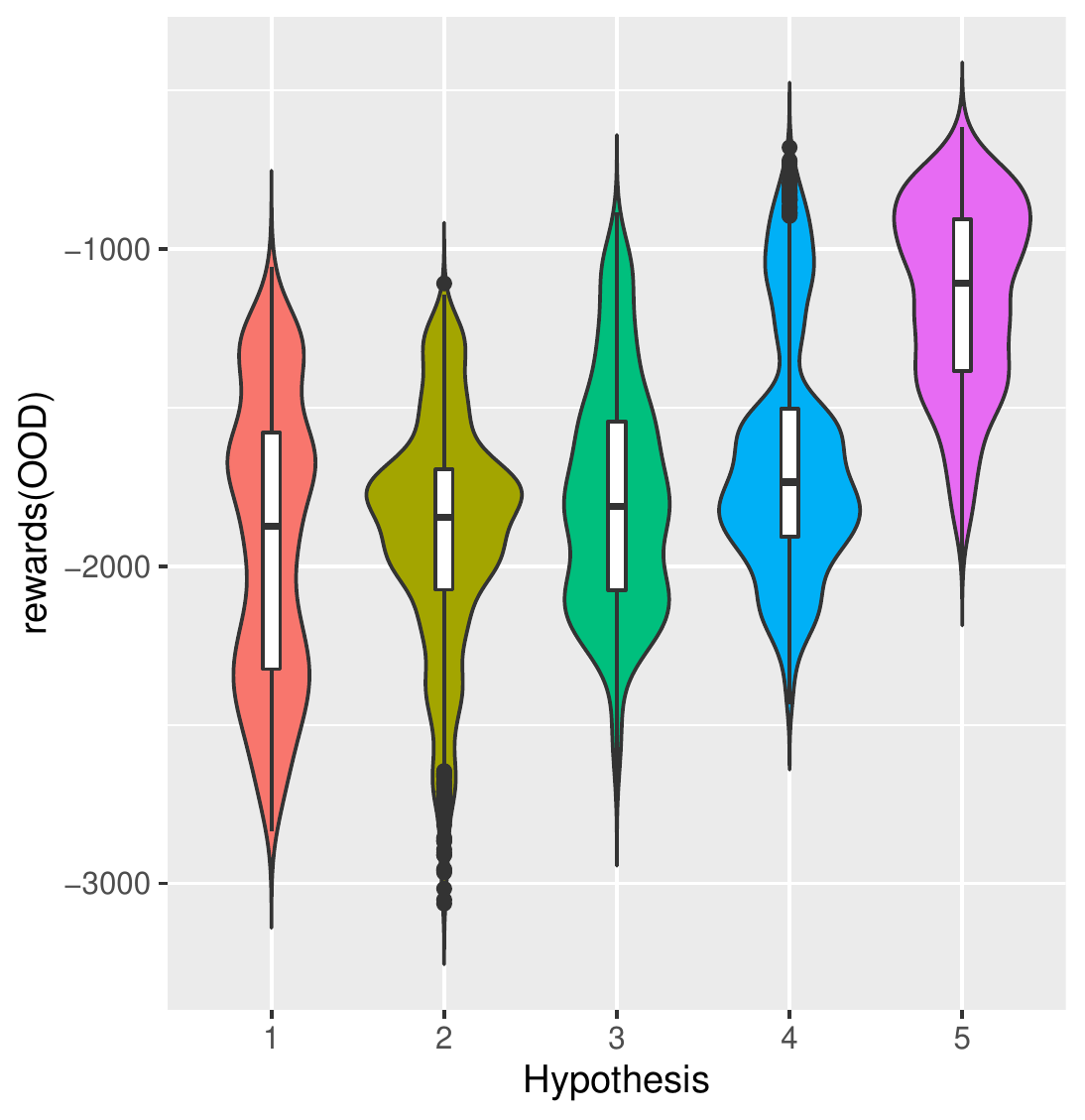}
    \includegraphics[width=0.24\linewidth]{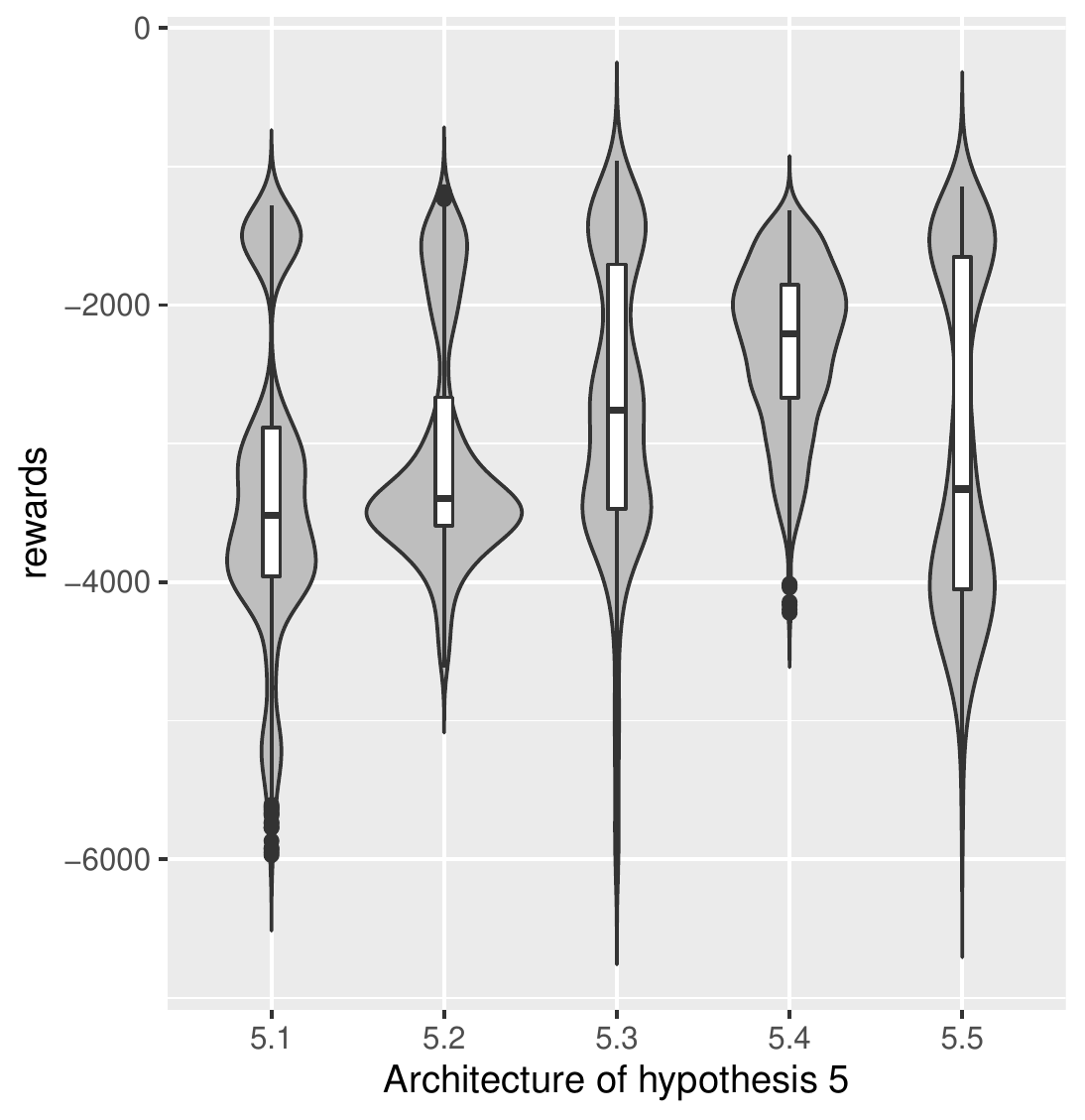}
    \includegraphics[width=0.24\linewidth]{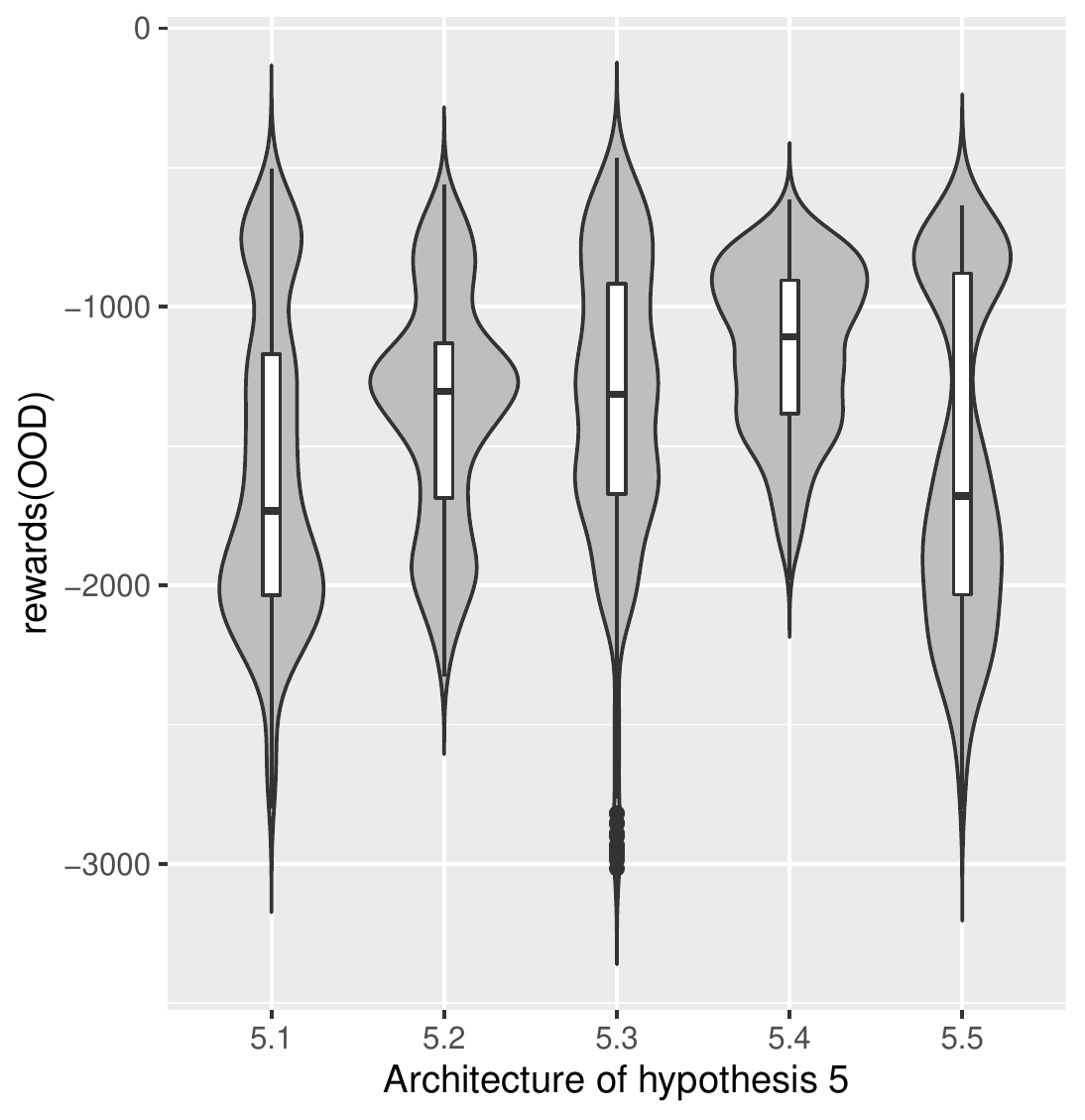}
    \caption{Comparison among the different hypotheses in "GhostRun" cooperative multi-agent reinforcement learning environment (preliminary results).
    We compared their rewards in the final 100 episodes (900-1000) in a testing environment that is the same as the training environment (IID) and a testing environment that is different from the training environment (OOD). The results are from 10 different random seeds. From left to right: (a) IID test rewards of main five hypotheses, (b) OOD test rewards for the five main hypotheses OOD.  (c) IID test rewards for five alternative architectures for Hypothesis 5 and (d) OOD test rewards for five alternative architectures for Hypothesis 5.}
    \label{fig:AmongHypotheses}
\end{figure}

\begin{figure*}
    \centering
    \includegraphics[width=0.24\linewidth]{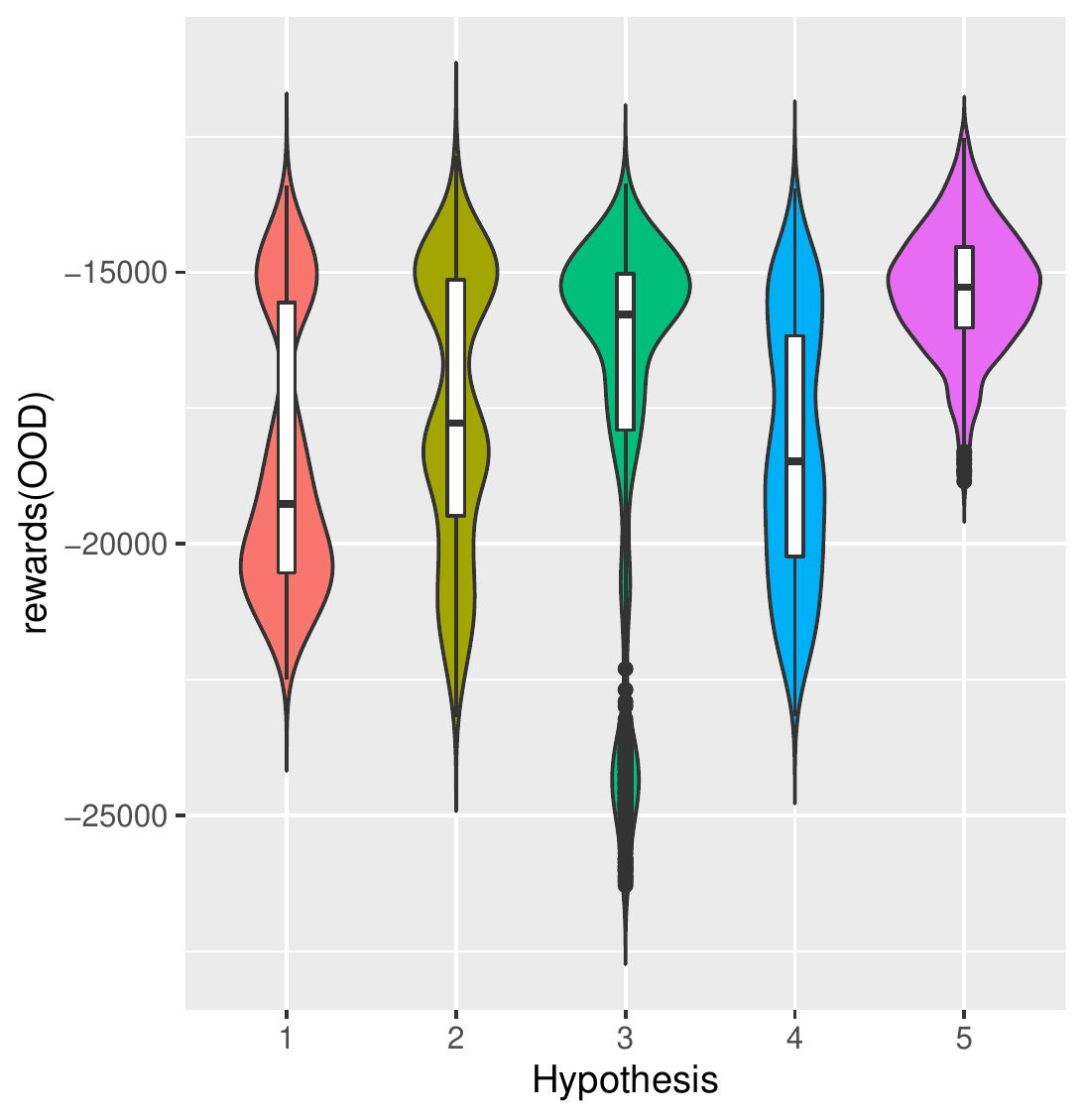}
    \includegraphics[width=0.24\linewidth]{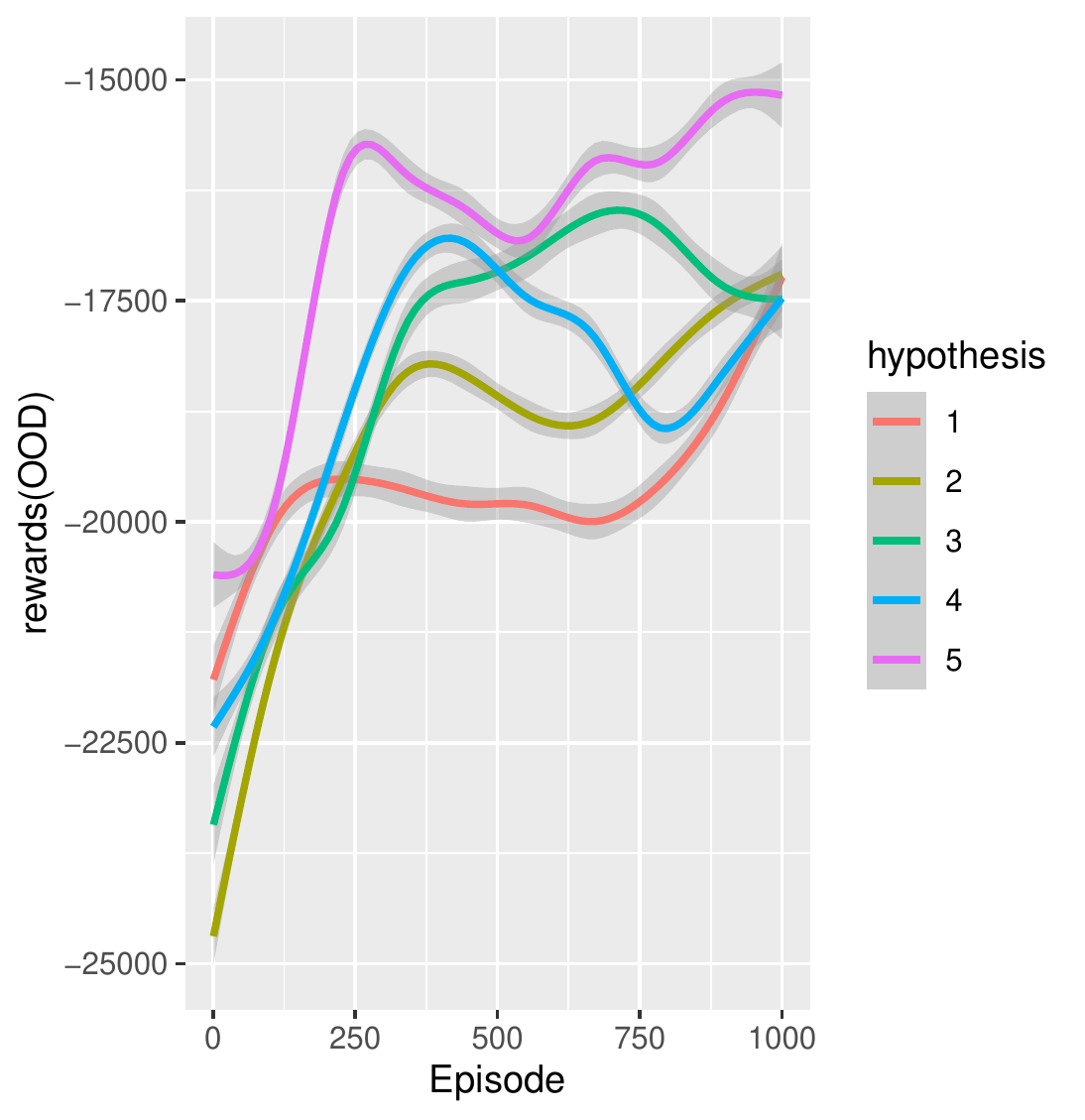}
     \includegraphics[width=0.24\linewidth]{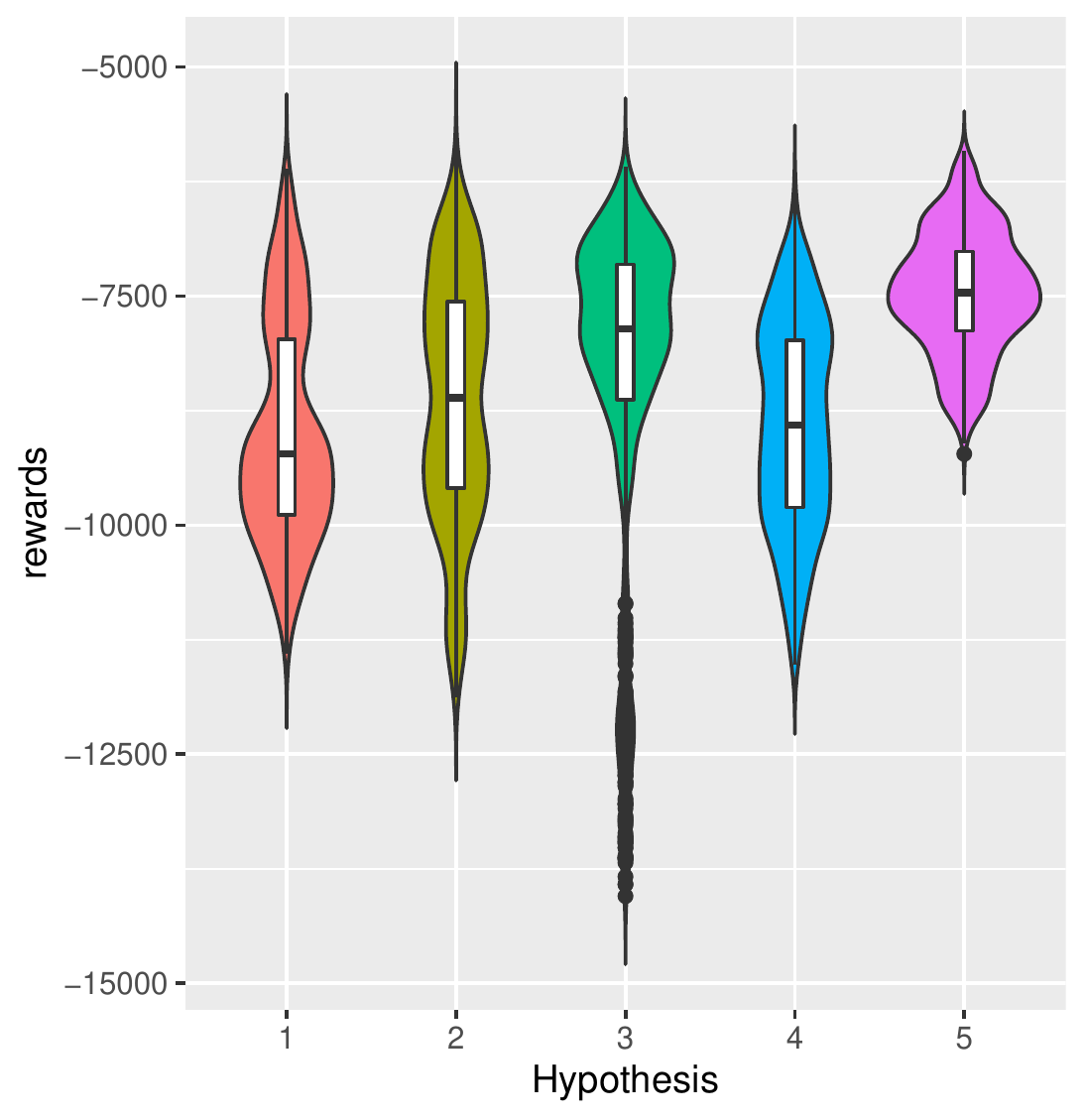}
    \includegraphics[width=0.24\linewidth]{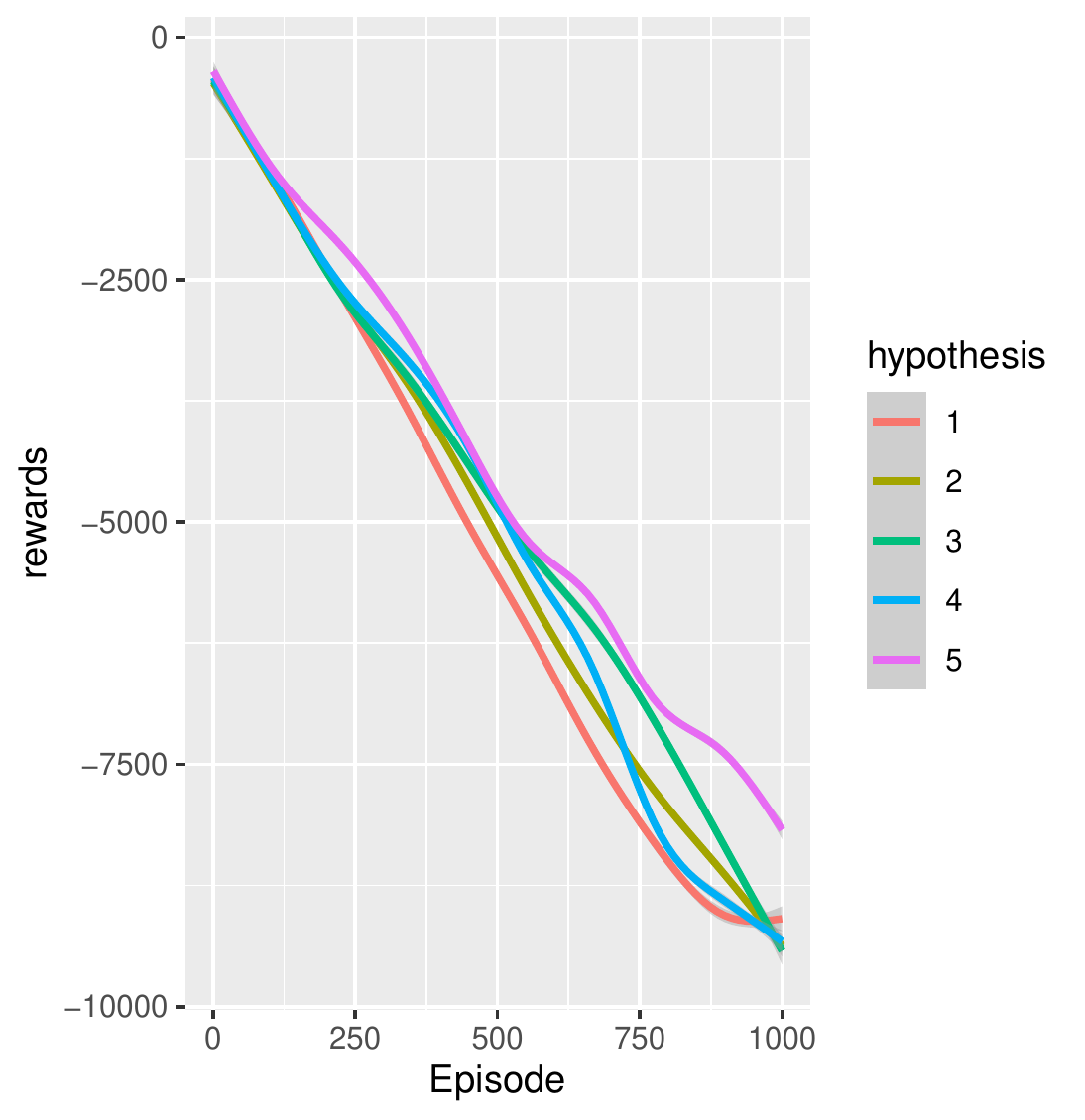}
    \caption{Preliminary results of performance of models in "GhostRun" corresponding to different hypotheses in a continual learning setting. From left to right: (a) reward distribution during learning, (b) timecourse of the learning reward across episodes, (c) reward during OOD evaluation, and (d) timecourse of the OOD test reward across episodes. Notice the common decrease in reward since the number of ghosts to escape from is increasing with episodes.}
    \label{fig:continual}
\end{figure*}

\begin{figure*}
    \centering
    \includegraphics[width=0.23\linewidth]{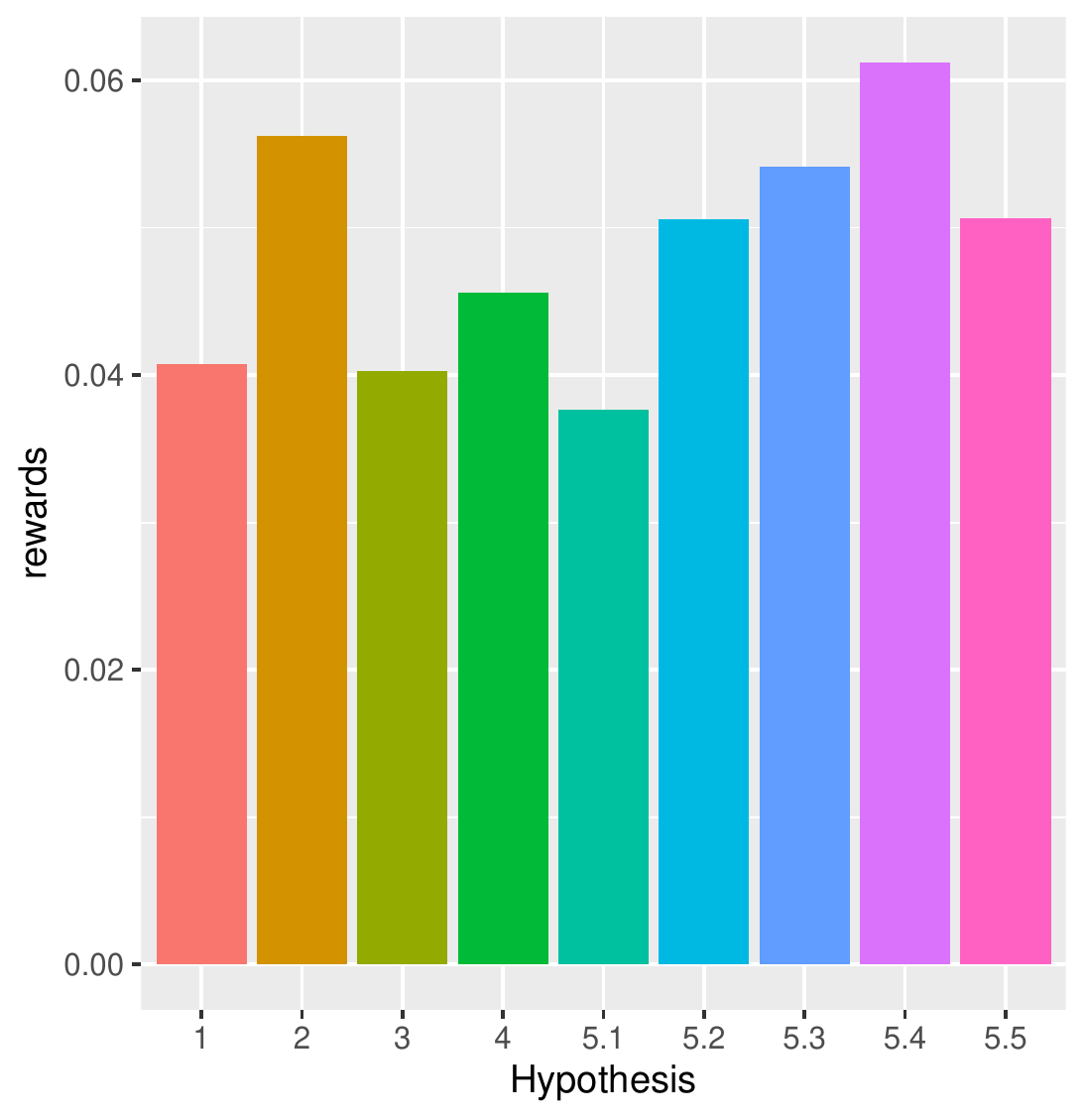}
    \includegraphics[width=0.23\linewidth]{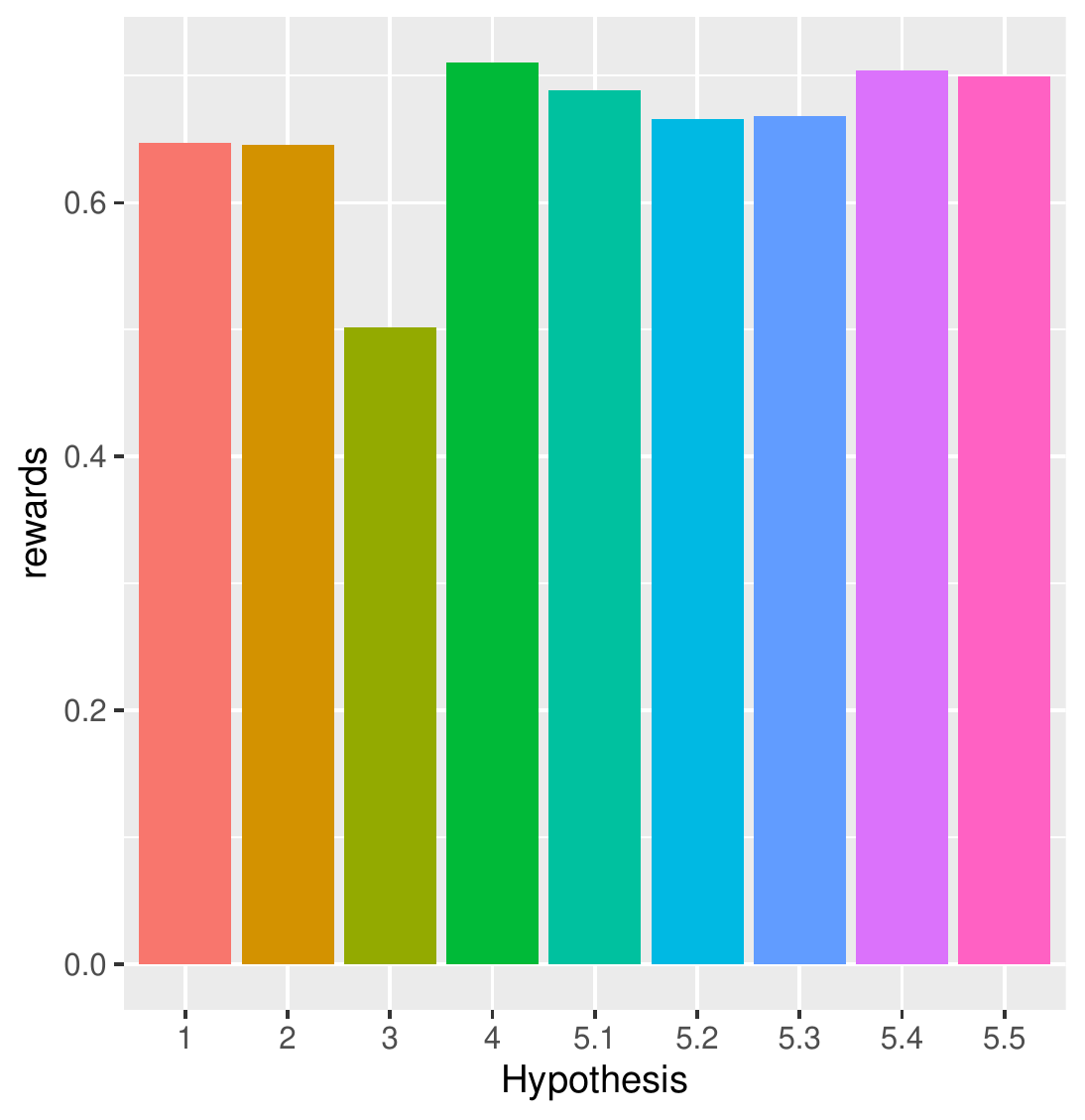}
    \includegraphics[width=0.23\linewidth]{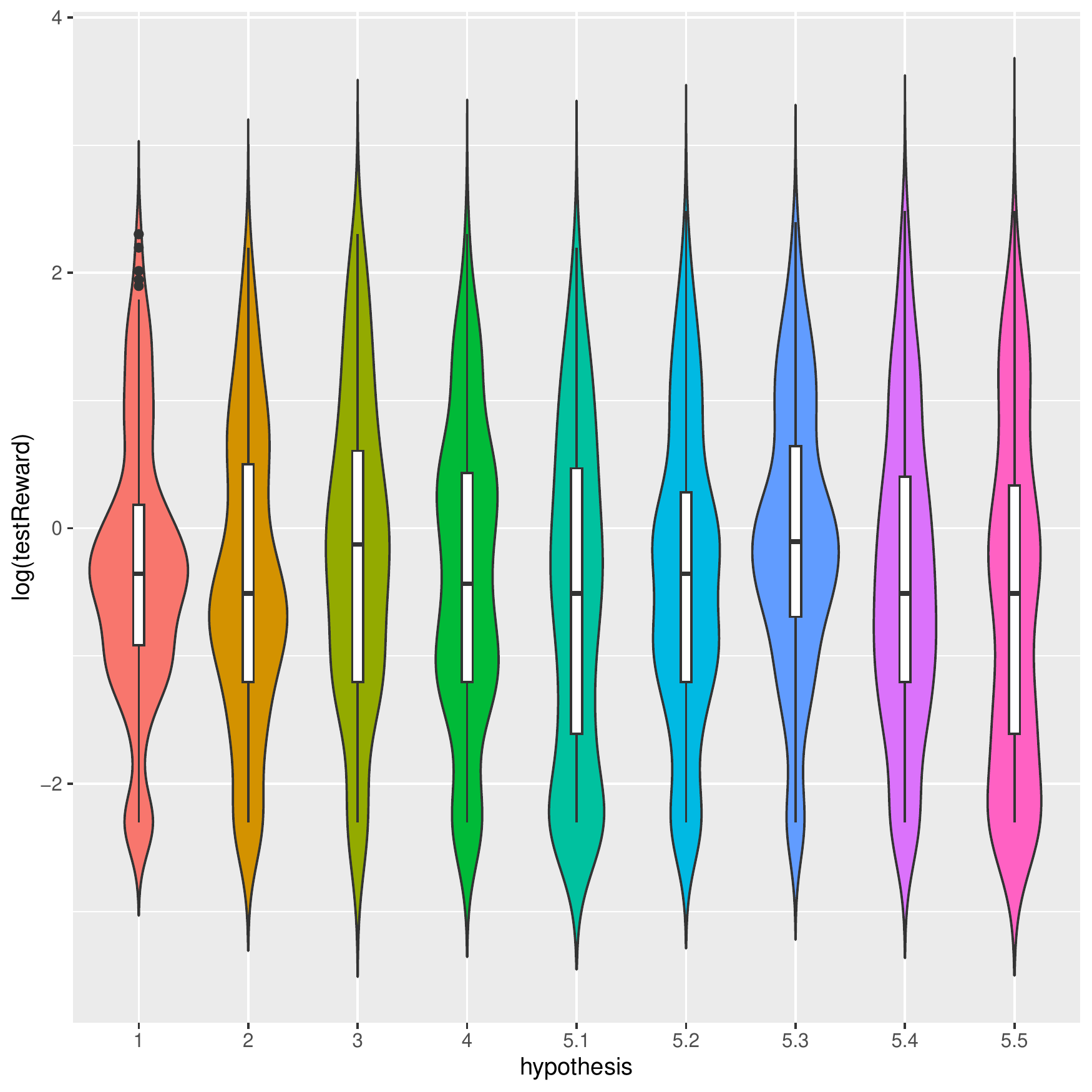}
     \includegraphics[width=0.23\linewidth]{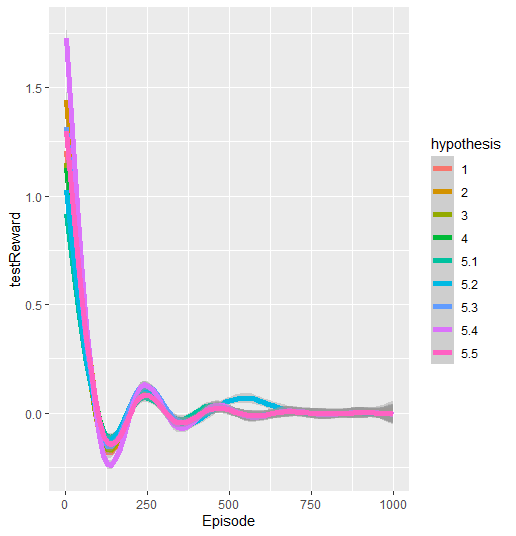}
    \caption{Preliminary results of performance of models in "MazeCleaners" corresponding to all the different hypotheses.  From left to right: (a) reward during learning, (b) reward during OOD evaluation, (c) test rewards distributions, and (d) timecourse of the test reward across episodes.}
    \label{fig:mazeResults}
\end{figure*}

\section{Experiments}
Experiments were conducted to investigate which of the five hypotheses show an advantage over others in MARL environments. 

\subsection{Comparison between the five main hypotheses}
In the first set of experiments, we compared the performance of the five architectures in the "GhostRun" MARL environments. The models obtained from training were tested both in an environment that is the same as the training setting (Independent and Identically Distributed or IID testing) and in an environment with distributional shifts (Out-Of-Distribution or OOD testing). Our results suggest that hypothesis five, corresponding to the attention schema theory, achieves the best performance (Figure \ref{fig:AmongHypotheses}). The results are from 10 different random seeds.

\subsection{Ability to generalize in a continuous learning setting}
We compared the performance of the different architectures in a continuous learning setting, in which the difficulty of the GhostRun task increases every 50 episodes --- with the addition of an additional ghost (Figure \ref{fig:continual}).

\subsection{Replication in different task and environment}
We replicated the main results of "GhostRun" in the "MazeCleaners" environment (Figure \ref{fig:mazeResults}). Here, Hypothesis 5.4 also gave the best performance among all architectures.



\section{Discussion and Future Work}

In this paper, we compared five different possible relationships between attention and its internal control, as suggested by the cognitive science literature (Figure 1) \cite{graziano2015attention}. We have shown, through a sequence of experiments performed in MARL environments, that equipping agents with an internal control that models their attention helps coordination across agents to achieve a common goal.

If our simple experiments support the usefulness of the attention schema for MARL, much more work is needed to investigate the role of internal control of attention in a variety of different environments and in relation to different algorithmic components. Shared central critics \cite{iqbal2019} and joint attention \cite{lee2021joint} have been shown to promote robust and scalable learning in complex environments. Our study proposes a decentralized way to go in the same direction. Many perspectives have now been opened up.

First, the experiments used tasks entirely within 2D environments, due to the level of control they afford. The long-term plan is to test those architectures in richer environments, from simple 3D environments to photorealistic video game engines, and finally with robots \cite{bolotta2022}. Such a gradual increase of complexity will facilitate the creation of agents able to interact with humans in the real world.

Second, we have implemented the internal control module as a traditional recurrent neural network. This incorporates many core ideas of how an attention schema is meant to work at the functional level, but we acknowledge that it is still far from being a rich, coherent, descriptive, and predictive model of attention.

Third, we did not explore how the internal control module could be beneficial for linguistic skills or other high-level cognitive functions. With the theory of mind, the agent generically attributes mental states to other agents, including their desires, beliefs, and intentions \cite{rabinowitz2018}. Although our results support the idea that building an internal model of attention could be helpful for social tasks, future work will test whether the control module is indeed capable of representing and predicting the attention of others.

The applications of this work are several. Learning a rich and schematic model of one's own attention will lead to better decision making, especially in contexts where there are many distractions and deep focus is hard to achieve \cite{stone2021}. Using an attention schema to infer the state of other agents' attention will also be useful for settings in which modeling opponents \cite{albrecht2017, he2016}, value alignment \cite{fisac2020, hadfield2016} and communication \cite{oroojlooy2021, foerster2016} play a key role.

\section{Acknowledgment}
This study was supported by the Institute for Data Valorization, Montreal (IVADO; CF00137433) and enabled in part by support provided by Calcul Québec (www.calculquebec.ca) and Digital Research Alliance of Canada (www.alliancecan.ca). G.D. was supported by the Fonds de recherche du Québec (FRQ; 285289), Natural Sciences and Engineering Research Council of Canada (NSERC; DGECR-2023-00089), and the CIFAR. 

We thank Dr Anirudh Goyal from DeepMind for insightful discussion and suggestions in experiments.  We also thank the team of the project Astound (EIC Pathfinder Challenge Awareness Inside - GA Number: 101071191) and Dr Alex Lamb from Microsoft  research for  stimulating discussions of potential connection between the AST and the control of the attention in Transformers 


\bibliography{example_paper,bibliography.bib}
\bibliographystyle{tom2023}

\end{document}